\documentclass{article}

\usepackage[dblblindworkshop,final]{neurips_2026}
\workshoptitle{SLMs for Agentic Systems, Paris, France, 2026}

\usepackage[utf8]{inputenc}
\usepackage[T1]{fontenc}
\usepackage{hyperref}
\hypersetup{pdfauthor={Pierre-Olivier Bonin},pdftitle={Error Diffusion Replaces Latent Weights: Training Binary-Weight Language Models on a Few Bits of Dithered State}}
\usepackage{url}
\usepackage{booktabs}
\usepackage{amsfonts}
\usepackage{amsmath,amssymb}
\usepackage{graphicx}
\usepackage{microtype}
\usepackage{xcolor}
\usepackage{algorithm}
\usepackage{algpseudocode}
\usepackage{multirow}
\usepackage{capt-of}

\newcommand{\ed}{ED}
\newcommand{\bwt}[1]{#1\,b/wt}
\newcommand{\adaptlatent}{$1.533 \pm 0.007$ (3 seeds, at the best of three Adam learning rates)}

\title{Error Diffusion Replaces Latent Weights:\\Training Binary-Weight Language Models on a Few Bits of Dithered State}

\author{%
  Pierre-Olivier Bonin\\
  Independent Researcher\\
  \texttt{pierre-olivier.bonin@hotmail.com}
}

\begin{document}

\maketitle

\begin{abstract}
Binary-weight language models store one bit per weight, but training them keeps a full-precision latent copy of each weight plus Adam state, about 96 extra bits per weight, so a deployed binary model cannot be adapted on the device that runs it. We replace the latent weight with a 2--8-bit residual buffer quantized by stochastic rounding, which cuts the training state per weight from 97 bits to 3--9, up to $19\times$ less. The resulting error-diffusion (\ed{}) optimizer moves the residual-carrying loop of image halftoning onto the weight-update path. Each buffer accumulates the unrealized part of every update and flips the weight when the accumulated value crosses a threshold, and it works only when its quantization grid has a level exactly at that threshold. Under this threshold-alignment rule, 4-bit buffers match 8-bit buffers on a binary multilayer perceptron benchmark, and the same rule lifts a different latent-free optimizer, Bop, from a 10-point deficit at 4 bits to full-precision accuracy. On character-level transformers with 97\% of parameters binary, \ed{} trains within 5--11\% of the latent-weight reference at a 2{,}000-step budget and within 1.7\% at convergence. On scaling ladders over enwik8 and wikitext-103, up to 19M parameters on enwik8 (70M with the wikitext-103 embeddings), the latent-free loss stays within $-2.3\%$ to $+2.9\%$ of the reference at all six sizes once a per-step cap on the buffer update removes a launch instability (\S\ref{sec:cap}); without the cap it trails by 10\% at the largest tokenized size. A deployed binary model continues training under \ed{} at one tenth of the latent training state, matching latent-weight continuation, and a fused CPU kernel completes the adaptation within a 275\,MB training-memory budget, which latent-weight training cannot. A 250-step probe sets the flip learning rate at any new scale.
\end{abstract}

\section{Introduction}

Small language models suit agentic systems because they run where the agent acts, on a phone, a laptop, or an embedded controller. Binary-weight networks push this furthest, with $32\times$ memory compression and multiplication-free inference \citep{arxiv1603_05279,arxiv2310_11453}. The standard training recipe since BinaryConnect \citep{arxiv1511_00363} keeps a full-precision (FP) \emph{latent} weight per binary weight, updates it with Adam \citep{kingma2015adam}, and binarizes it on the fly, as do XNOR-Net \citep{arxiv1603_05279}, DoReFa \citep{arxiv1606_06160}, BitNet \citep{arxiv2310_11453,arxiv2402_17764}, and binary normalized networks \citep{arxiv2509_07025}. The training state per weight is then 97 bits (a 32-bit latent weight, two 32-bit Adam moments, and the sign), so a 100M-weight binary model occupies 12.5\,MB at rest but needs about 1.2\,GB of training state to be adapted. Latent-free alternatives exist but keep at least 32 bits per weight. The binary optimizer Bop \citep{arxiv1906_02107} replaces the latent weight with a 32-bit exponential moving average (EMA) of gradients, and BOLD \citep{arxiv2405_16339} keeps an accumulator of unspecified precision.

We transfer a mechanism from image processing. A grayscale image survives 1-bit quantization because Floyd--Steinberg error diffusion \citep{floyd1976adaptive} carries each pixel's rounding residual forward until it accumulates into a pixel flip. Signal processing knows this loop as sigma-delta ($\Delta\Sigma$) modulation \citep{anastassiou1989ed}, error feedback (EF) in gradient compression \citep{seide2014onebit,karimireddy2019ef} applies it in time, and stochastic rounding (SR) \citep{hohfeld1992probabilistic} applies it in value. We apply it to the weight-update path, so the residual buffer replaces the latent weight, and we quantize the buffer to 2--8 bits with SR, which cuts the training state per weight from 97 bits to 3--9 (Appendix~\ref{app:concept}).

\paragraph{Contributions.}
\begin{enumerate}\setlength{\itemsep}{0pt}\setlength{\topsep}{2pt}
\item We propose \ed{}, whose entire per-weight training state is one sign bit plus a 2--8-bit SR-quantized error-feedback buffer, $19\times$ less than latent-weight training at 4 bits (\S\ref{sec:method}).
\item We show that the buffer is a stochastic accumulate-and-fire unit whose grid must contain a level at the flip threshold, a rule that removes an apparent 4-bit floor and restores full-precision accuracy to dithered Bop at 4 bits (\S\ref{sec:rule}).
\item We show that 97\%-binary transformers train within 5--11\% of the latent reference at 2{,}000 steps and 1.7\% at convergence, that the loss tracks the reference at all six ladder sizes once a per-step update cap removes a launch instability, and that a deployed model continues training on a CPU within a 275\,MB training-memory budget (\S\ref{sec:gpt}--\S\ref{sec:adapt}).
\item We give a 250-step probe that sets the flip learning rate at a new scale, validated by a pre-registered prediction (\S\ref{sec:drive}).
\end{enumerate}

\section{Related work}\label{sec:related}

\paragraph{Latent-free binary training.} Bop \citep{arxiv1906_02107} flips a weight when a 32-bit gradient EMA crosses a threshold and is the closest prior method; \citet{yayla2022dac} compress its state by about $2\times$. BOLD \citep{arxiv2405_16339} and the binary-error-signal line \citep{arxiv2412_00119,arxiv2512_04189} train without backpropagated gradients. None combines backpropagated updates with a low-bit SR-quantized residual as the sole per-weight state.

\paragraph{Error feedback.} 1-bit SGD, EF-SignSGD, and EF21 \citep{seide2014onebit,karimireddy2019ef,arxiv2106_05203} correct \emph{communication} compression while FP weights persist underneath. ECO \citep{arxiv2601_22101} removes master weights at FP8/INT4 by feeding the quantization error into Adam momentum, so \bwt{64} of optimizer state remains, and MicroAdam \citep{arxiv2405_15593} quantizes an error-feedback buffer beside FP weights. Sigma-delta modulation appears as activation coding \citep{arxiv1611_02024,arxiv1706_04159} and as post-training weight quantization \citep{arxiv2510_03275,arxiv2404_08131,arxiv2010_15979}. In all of these the modulator's output is a message or a post-hoc quantization, whereas in \ed{} it is the stored model.

\paragraph{Precision and scale.} Scaling-law studies of quantized training \citep{arxiv2411_04330,arxiv2502_16440,arxiv2505_14302} cover latent-weight training only, so \S\ref{sec:ladder} tests whether parallel loss curves survive removing the latent weight. $\mu$P \citep{arxiv2203_03466} and u-$\mu$P \citep{arxiv2407_17465} derive learning-rate transfer from continuous latent weights, which \ed{} lacks, and we are not aware of a cross-scale rule for Bop's threshold (\S\ref{sec:drive}).

\section{Method}\label{sec:method}

\paragraph{Notation.} Each binary weight $w \in \{-1,+1\}$ is scaled by a fixed per-layer scalar $s = \mathbb{E}|W_{\text{init}}|$. All quantities are in \emph{flip units}, where the distance between the two weight states is 2 and the flip threshold is 1. The proxy gradient $g$ comes from the straight-through estimator (STE) \citep{bengio2013ste}, and the flip learning rate $\eta$ (flip-lr) scales it into a demanded movement $u$. Each weight owns a buffer $e$ on a $2^b$-level uniform grid over $[-C, C]$, where $b$ is the buffer width in bits and $C$ the clamp, $\operatorname{SR}_b$ rounds stochastically onto that grid, and $F$ is the flip indicator.

\begin{algorithm}[t]
\caption{One \ed{} step for one binary layer (all quantities in flip units).}
\label{alg:ed}
\begin{algorithmic}[1]
\Require signs $w$, buffer $e$, proxy gradient $g$, flip-lr $\eta$, scale $s$, clamp $C$, bits $b$
\State $u \gets -\eta\, g / s$ \Comment{demanded movement}
\State $t \gets e + u$ \Comment{accumulate}
\State $F \gets \mathbf{1}\left[\operatorname{sign} t \neq \operatorname{sign} w \;\wedge\; |t| \ge 1\right]$ \Comment{fire}
\State $w \gets w\,(1 - 2F)$ \Comment{flip where $F = 1$}
\State $e \gets \operatorname{SR}_b\!\left(\operatorname{clamp}(t - 2F\operatorname{sign} t,\; -C,\; C)\right)$ \Comment{carry the residual, then quantize}
\end{algorithmic}
\end{algorithm}

\paragraph{The update.} Algorithm~\ref{alg:ed} gives one step. A flip fires when the accumulated value points away from the current sign and exceeds the threshold, and the realized movement (2 flip units) is subtracted before the remainder is rounded onto the buffer grid. No latent weight or optimizer moment exists. We read the loop as a first-order $\Delta\Sigma$ modulator of the update stream, with the flip as its 1-bit quantizer. Deterministic rounding in line~5 halts training at every $b$ (\S\ref{sec:rule}), so SR is the firing mechanism. Binary activations use $\operatorname{sign}$ with a per-batch abs-mean scale and a clipped-STE backward pass. The \textbf{threshold-alignment rule} of \S\ref{sec:rule} requires a grid level exactly at the flip threshold, that is $C = 1$.

\paragraph{Memory accounting.} Every reported cell carries bits of persistent training state per weight (b/wt). FP32+Adam carries 96, BitNet-style latent training 97, Bop 33, bf16 mixed precision 112, and \ed{} $1{+}b$ for $b\in\{2,\dots,8\}$ (\bwt{3}--\bwt{9}). The clamp adds an off-grid rail state ($2^b{+}1$ reachable values, 4.09 bits at $b{=}4$), which we round down to $b$ bits in favor of \ed{}. Per-layer weight scales, per-batch activation scales, and the FP embeddings, LayerNorms, and head of the transformers are excluded from the per-weight count (\S\ref{sec:setup}).

\section{Experimental setup}\label{sec:setup}

\paragraph{Tasks and data.} (i)~A teacher--student MLP benchmark isolates feature learning. Inputs are $x \sim \mathcal{N}(0, I_{64})$ and labels come from a fixed random 4-layer $\tanh$ teacher with 10 classes (20k train, 4k test). The student is a frozen random FP projection, two binary hidden layers of width $W$ with binary activations (the only feature learners), and a trainable FP readout, and the metric is test accuracy (higher is better). (ii)~Character-level Shakespeare \citep{karpathy2015charrnn} (1.1\,MB) trains nanoGPT-style \citep{karpathy2022nanogpt} 4-layer transformers with $d_{\text{model}} \in \{128, 256, 512\}$ for 2{,}000 steps, and $d_{\text{model}}{=}256$ for 50{,}000 steps. (iii)~A byte-level enwik8 ladder \citep{hutter2006prize} trains 0.5M, 3.3M, and 19M parameters (2, 4, and 6 layers) for 15k steps of $64 \times 256$ bytes, and (iv)~a tokenized ladder trains the same binary cores on wikitext-103 \citep{merity2017wikitext} with GPT-2 byte-pair encoding \citep{radford2019gpt2} (118M tokens, block 128). (v)~Deploy-then-adapt pretrains a latent-weight binary transformer (4 layers, $d{=}256$) on enwik8, discards the latent weights, and adapts the deployed artifact to byte-level Shakespeare for 5{,}000 steps. For (ii)--(v) the metric is the best validation loss in nats (lower is better). In every transformer the attention and MLP projections are binary (97\% of parameters at $d{=}256$ on characters), and the embeddings, LayerNorms, and head are FP and trained by AdamW \citep{loshchilov2019adamw}.

\paragraph{Baselines.} FP32+Adam (\bwt{96}) is the accuracy upper bound, BitNet-style latent-weight training (\bwt{97}, the ``latent reference'') is the standard recipe and shares the STE with \ed{}, and dithered Bop (\bwt{9} at 8 bits) is the closest latent-free method and runs on the MLP benchmark. A frozen-random-binary control (binary layers never updated) is the floor above which accuracy is attributable to the update rule (Appendix~\ref{app:extra} adds three further controls).

\paragraph{Protocol.}\label{sec:protocol} The flip-lr is the one hyperparameter of \ed{}. MLP cells use flip-lr 1.0 at every width. The 2{,}000-step Shakespeare cells use the best of a five-point sweep on seed 0, reused for the other seeds. Every longer transformer cell and every adaptation cell sets its flip-lr by a 250-step drive probe. We train 250 steps at two or three candidate flip-lrs, record the mean absolute buffer occupancy $\bar{|e|}$ at step 250, compute the drive $\rho = \bar{|e|}/250$ in flip units per step, and keep the candidate with $\rho \in [1.0, 1.3]\times10^{-3}$ (Appendix~\ref{app:probe}). Section~\ref{sec:drive} shows that sweep-selected and probe-selected flip-lrs lie in or near this band. FP and latent arms use AdamW at $3\times10^{-4}$ with no schedule unless stated. Runs of 15k steps and longer decay the flip-lr linearly to $0.1\times$. In the runs of \S\ref{sec:cap}, a stall check restarts a run with a new seed when its validation loss at step 2k exceeds its value at step 1k, and a divergence guard does the same when, after step 2k, the validation loss exceeds the run's best value by more than 0.3 nats; a cell is reported as diverged when three launches hit it. Runs of 15k steps and longer also stop early when the validation loss has not improved by 0.005 nats over five consecutive evaluations; the reported value is the best one. Before the transformer runs we fixed a pass criterion of $+20\%$ against the latent reference, and we report every cell whether or not it passes (Appendix~\ref{app:prereg} lists every pre-registered prediction). Every table states its seed count, and $\pm$ is the standard deviation over seeds (divisor $n$). All runs use one NVIDIA RTX~4080 SUPER (16\,GB), and Appendix~\ref{app:hparams} lists every hyperparameter and wall-clock time.

\section{Results}

\subsection{The buffer is a stochastic accumulate-and-fire unit}\label{sec:rule}

\begin{figure}[t]
\centering
\includegraphics[width=\linewidth]{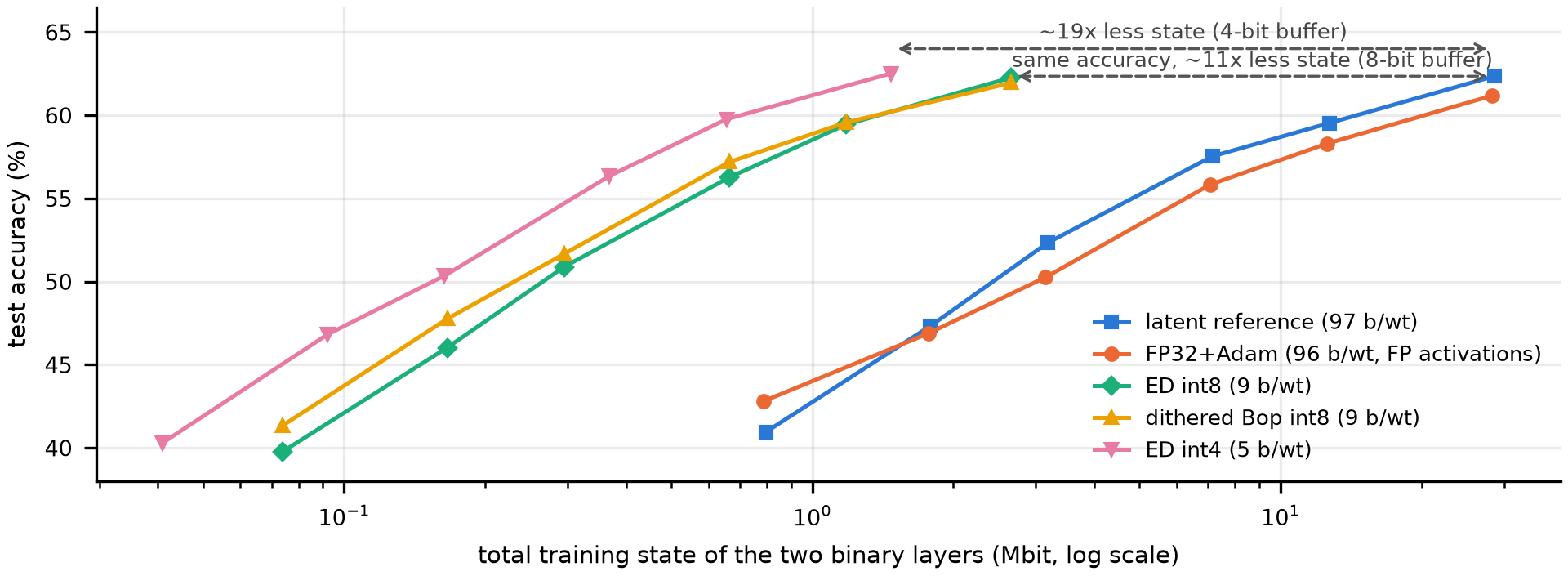}
\caption{MLP-benchmark test accuracy against the training state of the two binary layers ($2W^2 \times$ b/wt), widths 64--384, 5 seeds, $C{=}1.0$. At the accuracy of the latent reference, the 8-bit buffer needs $11\times$ less training state and the 4-bit buffer $19\times$ less.}
\label{fig:exchange}
\end{figure}

\textbf{Does a quantized buffer match a latent weight?} At width 192 (5 seeds), the frozen control scores $35.3 \pm 0.9$, the latent reference $57.5 \pm 1.0$ (\bwt{97}), \ed{} with an 8-bit buffer $56.3 \pm 1.4$ (\bwt{9}), and \ed{} with a 4-bit buffer $56.4 \pm 1.3$ (\bwt{5}). The latent reference's advantage decays with width and vanishes by 256--384 ($62.4 \pm 0.3$ vs.\ $62.3 \pm 0.3$ at 384). Figure~\ref{fig:exchange} plots accuracy against the total training state of the two binary layers (Appendix~\ref{app:extra}, Table~\ref{tab:exchange}). At equal accuracy the 4-bit buffer needs about $19\times$ less training state than the latent reference and the 8-bit buffer about $11\times$ less.

\begin{table}[t]
\centering
\caption{Test accuracy (\%, higher is better) at width 192, 5 seeds, for the \ed{} buffer and for dithered Bop's SR-quantized EMA (FP EMA $56.6 \pm 1.5$ at \bwt{33}), with and without a grid level at the threshold. Bold marks the better cell of a pair by more than one std. \ed{} cells use flip-lr 1.0 (2-bit rule cell: best of four flip-lrs, 52.3 at 0.5; 51.8 at 1.0); Bop cells take the best of three $\tau$.}
\label{tab:crule}
\footnotesize
\begin{tabular}{lcc@{\qquad}cc}
\toprule
& \multicolumn{2}{c}{\ed{} buffer} & \multicolumn{2}{c}{dithered Bop EMA} \\
\cmidrule(lr){2-3}\cmidrule(lr){4-5}
bits & naive $C{=}1.5$ & rule $C{=}1.0$ & absmax grid & aligned grid \\
\midrule
8 & $56.4 \pm 1.4$ & $56.3 \pm 1.4$ & $55.9 \pm 1.5$ & $56.1 \pm 0.6$ \\
4 & $53.2 \pm 1.3$ & $\mathbf{56.4 \pm 1.3}$ & $46.5 \pm 1.3$ & $\mathbf{57.0 \pm 1.0}$ \\
2 & $33.1 \pm 4.9$ & $\mathbf{52.3 \pm 1.2}$ & $38.6 \pm 2.1$ & $\mathbf{51.0 \pm 1.2}$ \\
\bottomrule
\end{tabular}
\end{table}

\textbf{Where must the grid sit?} With the naive clamp $C{=}1.5$, accuracy falls below 8 bits and collapses at 2 bits (Table~\ref{tab:crule}, left), which looks like a 4-bit floor. With the grid strictly inside the threshold ($C \in \{0.5, 0.75, 0.9, 0.95\}$), training is inert at every bit width (zero flips in 25 epochs for $C \le 0.9$), because per-step drives of about $10^{-3}$ flip units never reach the threshold from the top grid level, and with levels beyond it ($C{=}1.5$) 2-bit buffers fire 1.3M times per epoch and score below the frozen control. At $C{=}1.0$ every flip is an SR jump onto a grid level with $|\ell| \ge 1$. The buffer accumulates sub-threshold evidence and SR converts the excess into a firing probability, so the top grid level must sit exactly at the threshold. Under this rule 4-bit buffers equal 8-bit buffers and 2-bit buffers recover from collapse (Table~\ref{tab:crule}). The apparent 4-bit floor was clamp geometry.

\textbf{Does the rule transfer to Bop?} Dithered Bop at \bwt{9} matches \ed{} at every width (Figure~\ref{fig:exchange}), but at 4 bits on an absmax-scaled grid it loses 10 points (Table~\ref{tab:crule}, right). We predicted before the run that regridding the EMA so that a level sits exactly at its threshold $\tau$ would restore full-precision accuracy (Appendix~\ref{app:prereg}), and it does, with $57.0 \pm 1.0$ at \bwt{5} against $56.6 \pm 1.5$ for the FP EMA at \bwt{33}, $+8.5$ points at 3 bits (45.3 vs.\ 53.8, not shown) and $+12.4$ at 2. At 8 bits the two grids tie, as expected once some level falls within one per-step drive of the threshold anyway. Threshold alignment is therefore a property of thresholded quantized accumulators in general, and it holds on the transformer: on the 3.3M enwik8 cell, Bop's 4-bit EMA collapses on the absmax grid ($2.46 \pm 0.24$) and reaches $1.669 \pm 0.004$ on the aligned grid, below \ed{}-int4 at the same \bwt{5} (1.707), while its FP EMA ties \ed{}-int8 (1.524 vs.\ 1.516; Appendix~\ref{app:cap}).

\textbf{Is the buffer just a small integer latent weight?} A $(k{+}1)$-bit saturating integer latent weight with SR updates ties the buffer at every matched budget, and a symmetric integer grid satisfies $C{=}1$ by construction, so the buffer coordinates are the frame in which the design variable is visible rather than a separate method. Appendix~\ref{app:extra} (Table~\ref{tab:controls}) adds a sign counter, which wins at 8 bits but collapses at 2, and two budget-split ablations that do not beat it.

\subsection{Transformers at a fixed budget and at convergence}\label{sec:gpt}

\begin{table}[t]
\centering
\caption{Best validation loss (nats per character, lower is better) on character-level Shakespeare. 2k-step columns: FP activations, mean $\pm$ std over 3 seeds, one flip-lr per \ed{} cell (100, 60, 80 at $d$ 128, 256, 512). 50k-step column: binary activations, 1 seed; parentheses: constant flip-lr. Bold marks the lowest loss per column.}
\label{tab:gpt}
\footnotesize
\setlength{\tabcolsep}{5pt}
\begin{tabular}{lcccc}
\toprule
& \multicolumn{3}{c}{2k steps} & 50k steps \\
\cmidrule(lr){2-4}\cmidrule(lr){5-5}
method & $d{=}128$ & $d{=}256$ & $d{=}512$ & $d{=}256$ \\
\midrule
FP32+Adam (\bwt{96}) & $\mathbf{1.727 \pm 0.045}$ & $\mathbf{1.569 \pm 0.008}$ & $\mathbf{1.570 \pm 0.009}$ & 1.591 \\
latent reference (\bwt{97}) & $1.971 \pm 0.018$ & $1.712 \pm 0.008$ & $1.576 \pm 0.001$ & $\mathbf{1.556}$ \\
\ed{}-int8 (\bwt{9}) & $2.065 \pm 0.021$ & $1.851 \pm 0.020$ & $1.755 \pm 0.012$ & 1.583 (1.586) \\
\ed{}-int4 (\bwt{5}) & $2.076 \pm 0.030$ & $1.855 \pm 0.042$ & $1.808 \pm 0.047$ & 1.681 \\
\bottomrule
\end{tabular}
\vspace{6pt}
\captionof{table}{Scaling ladders, best validation loss (lower is better), binary activations, probe-set flip-lrs, for the latent reference (\bwt{97}), \ed{}-int8 (\bwt{9}), and \ed{}-int4 (\bwt{5}), each size under two recipes: ``uncapped'' and ``cap'' (the per-step update cap of \S\ref{sec:cap}). enwik8 (nats per byte): 3 seeds, mean $\pm$ std, where a $\pm$ is shown, else 1 seed; the uncapped 19M \ed{}-int8 cell averages the 3 of 5 launches that did not stall (all 5: $1.357 \pm 0.085$). wikitext-103 (nats per token): 1 seed, except the capped 19M \ed{}-int8 cell at 15k (3 seeds); totals with the 50k-token embeddings: 13.3M, 28.9M, 70.4M. ``diverged'': all three launches hit the divergence guard (\S\ref{sec:protocol}). $\Delta$: \ed{} relative to the latent reference; the last column: int4 relative to int8 under the same recipe; bold is the lowest loss per size.}
\label{tab:ladder}
\scriptsize
\setlength{\tabcolsep}{3.5pt}
\begin{tabular}{lllccrcr}
\toprule
corpus & size & recipe & latent & \ed{}-int8 & $\Delta$ & \ed{}-int4 & $\Delta$ vs int8 \\
\midrule
enwik8 & 0.5M & uncapped & $1.943 \pm 0.007$ & $\mathbf{1.894 \pm 0.002}$ & $-2.5\%$ & $2.002 \pm 0.003$ & $+5.7\%$ \\
 &  & cap &  & $1.898$ & $-2.3\%$ & $2.007$ & $+5.7\%$ \\
enwik8 & 3.3M & uncapped & $\mathbf{1.468 \pm 0.006}$ & $1.516 \pm 0.012$ & $+3.3\%$ & $1.707 \pm 0.003$ & $+12.6\%$ \\
 &  & cap &  & $1.499$ & $+2.1\%$ & $1.711$ & $+14.1\%$ \\
enwik8 & 19M & uncapped & $\mathbf{1.239 \pm 0.002}$ & $1.288 \pm 0.002$ & $+3.9\%$ & $1.510 \pm 0.003$ & $+17.2\%$ \\
 &  & cap &  & $1.275 \pm 0.007$ & $+2.9\%$ & $1.509$ & $+18.3\%$ \\
\midrule
wikitext-103 & 0.5M & uncapped & $\mathbf{5.113}$ & $5.118$ & $+0.1\%$ & $5.224$ & $+2.1\%$ \\
 &  & cap &  & $5.114$ & $+0.0\%$ & $5.275$ & $+3.2\%$ \\
wikitext-103 & 3.3M & uncapped & $4.789$ & $4.782$ & $-0.1\%$ & diverged & -- \\
 &  & cap &  & $\mathbf{4.775}$ & $-0.3\%$ & $4.974$ & $+4.2\%$ \\
wikitext-103 & 19M, 15k & uncapped & $\mathbf{4.464}$ & $4.920$ & $+10.2\%$ & $4.749$ & $-3.5\%$ \\
 &  & cap &  & $4.510 \pm 0.006$ & $+1.0\%$ & $4.742$ & $+5.1\%$ \\
wikitext-103 & 19M, 30k & uncapped & $\mathbf{4.358}$ & $5.075$ & $+16.5\%$ & diverged & -- \\
 &  & cap &  & $4.371$ & $+0.3\%$ & $4.681$ & $+7.1\%$ \\
\bottomrule
\end{tabular}
\end{table}

\textbf{Does the latent-free transformer meet the criterion?} Table~\ref{tab:gpt} shows the 2{,}000-step cells. \ed{} with an 8-bit buffer trails the latent reference by 4.8\%, 8.1\%, and 11.4\% at $d_{\text{model}}$ 128, 256, and 512, inside the $+20\%$ criterion at every width, and the 4-bit buffer is within 0.06 of the 8-bit one. \textbf{Does the gap close with training?} In the 50{,}000-step regime (Table~\ref{tab:gpt}, last column) \ed{} reaches 1.583 with a linear flip-lr decay, within 1.7\% of the latent reference (1.556), and 1.586 without it. These single-run cells land within 0.035 of each other and FP32+Adam overfits this 1.1\,MB corpus well before 50k steps, so we do not rank them.

\subsection{Scaling ladders}\label{sec:ladder}

\textbf{Does the gap grow with scale?} Table~\ref{tab:ladder} gives both ladders. On byte-level enwik8, \ed{} with an 8-bit buffer beats the latent reference by 2.5\% at 0.5M parameters and trails it by 3.3\% at 3.3M and 3.9\% at 19M, so the gap stops widening between the two larger sizes. On tokenized wikitext-103 the two arms land within 0.15\% of each other at the two smaller sizes, but at 19M \ed{} trails by 10.2\% at equal budget and 16.5\% at a doubled budget. On this vocabulary the full-precision (FP) remnant, the parameters outside the binary layers (embeddings and head, 51.5M parameters) trained by Adam, outweighs the binary core 2.7:1, a regime the byte-level ladder never enters. \textbf{Is training stable?} At 19M on enwik8, two of five \ed{} launches stall in the first 2{,}000 steps and finish 0.14--0.20 worse, so the 19M cell averages the three runs that did not stall (9.5\% with the stalled runs included). Section~\ref{sec:cap} traces both the stalls and the tokenized gap to one event and removes it. \textbf{Does a 4-bit buffer scale?} At 2{,}000 steps it matches the 8-bit buffer (Table~\ref{tab:gpt}), but the gap opens with horizon and scale. On the enwik8 ladder, at probe-set flip-lrs two to four times below the 8-bit ones, the 4-bit buffer completes every size without stalls but trails the 8-bit buffer by 5.7\%, 12.6\%, and 17.2\%, and the 50k convergence cell lands at 1.681 against 1.583.

\subsection{Launch stalls are flip avalanches, and a per-step update cap removes them}\label{sec:cap}

Three questions remain from \S\ref{sec:ladder}: what causes the 19M stalls, and whether a buffer initialization or a flip-lr warm-up prevents them; whether the tokenized gap comes from a learning-rate mismatch between the AdamW-trained FP remnant and the \ed{} core; and whether the 4-bit buffer saturates over long horizons. Appendix~\ref{app:cap} holds the diagnostics behind this section.

\textbf{The stall.} With evaluations every 100 steps and per-block diagnostics, every stalled launch shows the same event. Inside a single 100-step window the flip rate doubles (38k to 79k flips per step on one captured event), the mean pre-activation magnitude of the first block rises 2.6-fold (1.28 to 3.39), the gradient norm reaching the binary weights falls five-fold (0.0097 to 0.0018), and the validation loss rises by 0.6--1.4 nats; LayerNorm gains, the head norm, and the FP remnant's gradient stay on their trend. The core then runs at a tenth of its flip rate and recovers slowly. The ``failure to ramp up'' of \S\ref{sec:ladder} is the aftermath of one flip avalanche in the first block; it is a property of the realization, not the seed, and neither a non-zero buffer initialization (5 of 20 launches failed, against 3 of 20, over two realizations of seeds 0--9) nor a warm-up prevents it. On a step-2000 checkpoint the magnitude of the per-weight update $u = -\eta\, g/s$ never exceeds 0.19 flip units on an ordinary step (median 0.0066), so a cap at 0.25 flip units, $u \leftarrow \mathrm{clamp}(u, -0.25, 0.25)$ before the buffer add, leaves ordinary steps untouched, removes the one-step route to an avalanche, and replaces the stall check of \S\ref{sec:protocol}.

\textbf{Results.} Table~\ref{tab:ladder}, capped column: where no avalanche occurred (the two smaller sizes on both corpora) the cap moves the loss by at most 0.007 nats in either direction (same seed). At 19M it removes the stalls (avalanche-free launches 16 of 20 $\to$ 13 of 13 on enwik8; completed launches 3 of 15 $\to$ 5 of 5 on wikitext-103) and lowers the loss: enwik8 $1.275 \pm 0.007$ against $1.288 \pm 0.002$ for the clean uncapped runs; wikitext-103 $4.510 \pm 0.006$ against $4.920$ at 15k steps and $4.371$ against $5.075$ at 30k. Under one recipe (probe-set flip-lr, linear decay, $C = 1.0$, binary activations, cap 0.25) the byte ladder reads $-2.3\%$, $+2.1\%$, $+2.9\%$ and the tokenized ladder $0.0\%$, $-0.3\%$, $+1.0\%$ ($+0.3\%$ at 30k) against the latent reference. The 19M tokenized gap was an avalanche-rate problem, not a capacity or learning-rate one: with embeddings and head frozen, \ed{} and the latent reference tie (7.279 against 7.243), and at a matched remnant learning rate of $10^{-4}$ they tie again (4.610 against 4.625). A fast remnant on a 50k-token vocabulary triggers the avalanche more often than byte-level data does; the remnant's size (2.7$\times$ the core at 19M) is unchanged and remains a limitation.

\textbf{The 4-bit buffer.} Histograms of the buffer over its grid, logged every 1k steps in replicates of the 50k-step cells of Table~\ref{tab:gpt}, decide the third question: the 4-bit buffer holds 6.2--7.5\% of its weights at every interior level from step 2k to 50k (uniform: 6.7\%) with at most 0.6\% on the rails, while the 8-bit buffer's mass gathers near the flip threshold (18.5\% in each outermost sixteenth of the range at 50k, 0.2\% on each rail). The 4-bit quantum ($2/15$ flip units) is about twenty times the median per-step drift, so rounding noise erases the accumulated signal: the gap is variance, not saturation, and it grows with horizon and scale because both shrink the drift relative to a fixed quantum. The wikitext-103 4-bit cells of Table~\ref{tab:ladder} were run under the same protocol: uncapped, 5 of 6 launches at the two smaller sizes and all three 30k launches diverged with a steady flip rate, a failure we have not diagnosed and that the cap does not remove (one of two capped 30k launches diverged as well); the completed capped runs trail the capped 8-bit arm by 3.2\%, 4.2\%, 5.1\%, and 7.1\% at 30k, against 5.7\%, 12.6\%, and 17.2\% on enwik8, where the cap changes the 4-bit cells by at most 0.005 nats.

\subsection{Deploy-then-adapt}\label{sec:adapt}

\textbf{Can a deployed binary model keep learning?} No latent copy exists, so the object being trained is the deployed binary network at every step, and adapting it in place adds only the $b$-bit buffer per weight. Loss falls from 2.32 to $1.520 \pm 0.003$ in 5{,}000 steps at \bwt{9} (3 seeds), against 1.95 for \ed{} from scratch at the same budget and \adaptlatent{} for continued latent-weight training with its full 97-bit state re-instantiated, so the buffer matches the re-instantiated latent weight. With a 4-bit buffer (\bwt{5}) adaptation reaches $1.610 \pm 0.005$, a 6\% cost. \textbf{Does it fit on inference-class hardware?} We ported the adaptation step to a fused CPU kernel and ran it on the 19M-parameter enwik8 model, adapting to Shakespeare bytes for 2{,}000 steps (batch 8, block 128, one seed). The deployed artifact packs eight signs per byte into 4.0\,MB, against 77\,MB for the float32 latents and FP remnant. The kernel applies the integer update during backpropagation, stores no weight gradients, and draws SR bits from a stateless counter-based hash \citep{salmon2011philox}. We poll resident memory every step and abort a run when its peak above the post-setup baseline (about 750\,MB of Python and PyTorch) exceeds a cap. Under a 275\,MB cap the fused kernel peaks at 254\,MB and reaches 1.513, within 0.017 of unconstrained latent-weight continuation (1.496), which peaks at 325\,MB and aborts under 150 and 200\,MB caps. The FP remnant still needs Adam state and the activation memory is unchanged.

\subsection{The constant-drive transfer rule}\label{sec:drive}

\textbf{Does one number set the flip-lr at every scale?} Across three corpora, widths 128--512, two activation regimes, 8-bit and 4-bit buffers, and warm-started adaptation, the selected flip-lr varies from 12 to 225, but the drive $\rho$ at each selected flip-lr lies within $0.85$--$1.73\times10^{-3}$ flip units per step, and off-band candidates lose (Appendix~\ref{app:extra}, Figure~\ref{fig:drive}). A pre-registered prediction at $d{=}512$ (optimal flip-lr 80--90, Appendix~\ref{app:prereg}) held, with 80 winning at loss 1.782 (binary activations, 1 seed), 160 tying, and 40 losing. The MLP harness does not log the step-250 occupancy, so the MLP cells are absent from the figure, and the rule does not extend through heavy-ball momentum. Unlike $\mu$P-style transfer \citep{arxiv2402_17457,arxiv2203_03466}, the rule is measured, not derived.

\section{Limitations}\label{sec:limitations}

The tasks are a synthetic MLP benchmark and language models with at most 19M parameters (70M with the wikitext-103 embeddings) and six layers on one GPU, evaluated by validation loss and synthetic accuracy, not downstream tasks. The 50{,}000-step and tokenized-ladder cells are single runs, so the small margins among the converged methods are not statistically separated. We do not claim wall-clock efficiency, the cost of generating SR bits is not measured, and every ladder size trains on fixed data, so the data dimension is untested \citep{arxiv2411_04330}. The FP remnant (embeddings, norms, head, and their Adam state) is now the dominant residual training memory. The 1-bit-gradient ablation (Appendix~\ref{app:extra}) is scoped to the weight-gradient operand and shows that gradient bits and buffer bits should not both be cut aggressively. We claim no accuracy advantage over a well-built saturating integer latent weight (\S\ref{sec:rule}).

\section{Conclusion}

\ed{} trains binary-weight MLPs and transformers on 3--9 bits of training state per weight, up to $19\times$ less than latent-weight training, by carrying each weight's unrealized update in a stochastically rounded buffer whose grid has a level at the flip threshold. With a per-step cap on the buffer update, the loss stays within $-2.3\%$ to $+2.9\%$ of the latent reference at all six ladder sizes (\S\ref{sec:cap}), and a deployed binary model continues training on a CPU within a 275\,MB training-memory budget. The FP remnant is the next target.

\section*{Acknowledgments}
We thank the two anonymous reviewers, whose questions on the 19M stalls, the full-precision remnant, and the 4-bit buffer led to \S\ref{sec:cap}.

\bibliographystyle{plainnat}
\bibliography{references_20260828}

\appendix

\section{Error diffusion in halftoning and in the weight update}\label{app:concept}

\begin{figure}[h]
\centering
\includegraphics[width=0.8\linewidth]{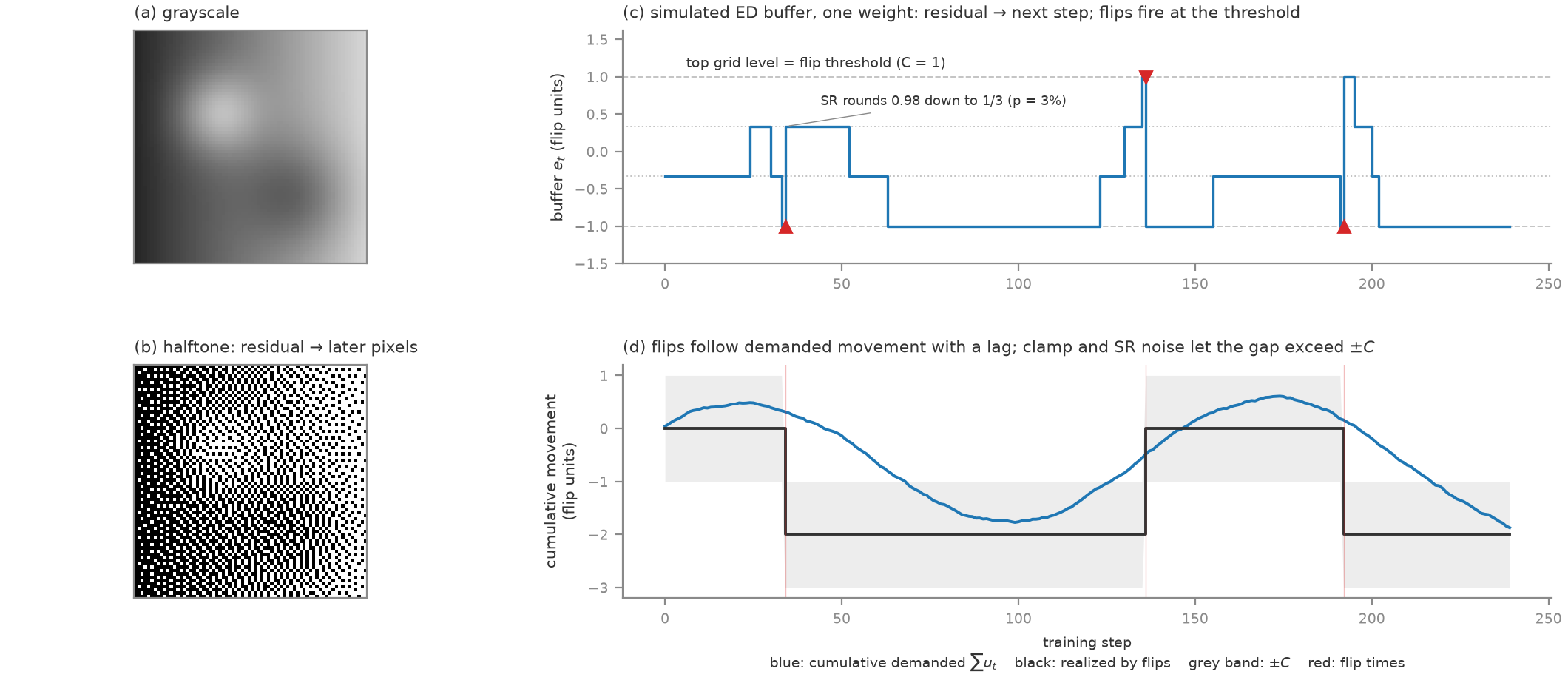}
\caption{Error diffusion in halftoning and in the weight update. Panels (a) and (b) show Floyd--Steinberg halftoning, which renders a grayscale image with binary pixels by carrying each pixel's rounding residual forward in scan order. Panel (c) shows the \ed{} optimizer applying the same recurrence in time, on a simulated weight with an illustrative drive. A binary weight's unrealized updates accumulate in a small quantized buffer (2-bit here), and a flip fires when stochastic rounding lands the buffer on the threshold level (\S\ref{sec:rule}). Panel (d) compares the cumulative demanded movement with the movement realized by flips. The flips follow the demand with a lag. The gap can exceed $C$, because the clamp at $\pm C$ discards updates that push a weight further toward its current sign, and stochastic rounding adds zero-mean noise that the buffer does not carry forward.}
\label{fig:concept}
\end{figure}

\section{The 250-step drive-probe protocol}\label{app:probe}

\begin{enumerate}
\item Configure the target run exactly (model, data, batch, block, binarization). Pick two or three candidate flip-lrs that bracket a linear-scaling guess from the nearest known scale.
\item Train 250 steps per candidate. Record the mean absolute buffer occupancy $\bar{|e|}$ at step 250 (flip units).
\item Compute the drive $\rho = \bar{|e|}/250$ in flip units per step. Select the candidate with $\rho \in [1.0, 1.3] \times 10^{-3}$. If none lands in the band, interpolate and re-probe.
\item The rule has known boundaries. It does not extend through heavy-ball momentum. At the largest widths prefer the band's lower edge, because at 4 bits an in-band probe reading can precede buffer saturation. Re-probe after any change of dataset, tokenizer, block size, or warm-start status.
\item The cost is the number of candidates $\times$ 250 steps (up to about 11 GPU-minutes per candidate at 19M parameters). For long runs add the stall check, which restarts with a new seed if validation loss at step 2k exceeds its value at step 1k.
\end{enumerate}

\section{Hyperparameters and compute}\label{app:hparams}

\paragraph{Transformers.} All transformer cells share one nanoGPT-style architecture (token and position embeddings, pre-LayerNorm blocks, causal attention, GELU MLP with $4\times$ expansion, untied bias-free head) and the settings in Table~\ref{tab:hparams}. The four projections of every block (query-key-value, attention output, MLP input, MLP output) are binary. The FP parameters (embeddings, LayerNorms, head) and, in the latent arms, the latent weights are trained by AdamW with learning rate $3\times10^{-4}$, $\beta = (0.9, 0.95)$, weight decay 0.01, no warmup, no schedule, no gradient clipping, and dropout 0. Weights initialize as $\mathcal{N}(0,1)/\sqrt{d_{\text{in}}}$ and the per-layer scale $s = \mathbb{E}|W_{\text{init}}|$ is fixed. Binary activations apply $\operatorname{sign}(z/s)\,s$ with a per-batch abs-mean scale $s$ to the MLP hidden layer only, with a clipped STE that passes the gradient where $|z/s| \le 1$, and attention internals stay FP. Validation loss is the mean over a fixed set of validation batches (50 batches of 64 sequences, 25 for wikitext-103) drawn with a fixed generator, and every cell reports the best validation loss over its evaluations. One CPU generator per run, seeded by the seed, drives initialization, batch sampling, and SR draws. The flip-lr decays linearly to $0.1\times$ in runs of 15k steps and longer.

\begin{table}[h]
\centering
\caption{Transformer cells. ``Layers/heads/$d$'' give depth, head count, and width; ``binary'' is the number of binary weights and ``FP'' the FP remnant. Flip-lrs are for the 8-bit buffer with the 4-bit value in parentheses where a 4-bit cell exists. The deploy-then-adapt latent arm was swept over Adam learning rates $\{10^{-4}, 3{\times}10^{-4}, 10^{-3}\}$.}
\label{tab:hparams}
\scriptsize
\setlength{\tabcolsep}{3pt}
\begin{tabular}{p{1.5cm}p{1.8cm}p{2.3cm}p{2.5cm}p{0.65cm}p{2.0cm}p{1.15cm}}
\toprule
cell & layers/heads/$d$ & binary / FP params & data, block, batch, steps (eval every) & binact & flip-lr & seeds \\
\midrule
Table~\ref{tab:gpt}, 2k & 4/4/128, 256, 512 & 0.79M/52k, 3.1M/103k, 12.6M/207k & Shakespeare char, 256, 64, 2k (250) & no & 100, 60, 80 (100, 60, 80) & 0, 1, 2 \\
Table~\ref{tab:gpt}, 50k & 4/4/256 & 3.1M/103k & Shakespeare char, 256, 64, 50k (1000) & yes & 60 (45) & 0 \\
enwik8 ladder & 2/4/128, 4/4/256, 6/8/512 & 0.39M/100k, 3.1M/201k, 18.9M/407k & enwik8 bytes, 256, 64, 15k (1000) & yes & 25, 60, 180 (12, 30, 45) & 0, 1, 2 (+3, 4 at 512) \\
wikitext-103 ladder & same three & 0.39M/12.9M, 3.1M/25.8M, 18.9M/51.5M & GPT-2 BPE, 128, 64, 15k (30k at 512) (1000) & yes & 40, 90, 225 & 0 \\
deploy-then-adapt & 4/4/256 & 3.1M/201k & Shakespeare bytes, 256, 64, 5k (250) & yes & 60 (30) & 0, 1, 2 \\
1-bit gradients & 4/4/256 & 3.1M/103k & Shakespeare char, 256, 64, 2k (250) & yes & adaptive drive 0.03 & 0, 1, 2 \\
CPU kernel & 6/8/512 & 18.9M/341k & Shakespeare bytes, 128, 8, 2k (500) & yes & 60 & 0 \\
\bottomrule
\end{tabular}
\end{table}

\paragraph{MLP benchmark.} The teacher draws 24{,}000 inputs $x \sim \mathcal{N}(0, I_{64})$ with a fixed seed and labels them by the argmax of a 4-layer $\tanh$ network (64$\to$48, then three 48$\to$48 layers) followed by a random $48\times10$ readout, and the first 20{,}000 examples train while the last 4{,}000 test. The student is a frozen random FP projection $64 \to W$ with ReLU, two binary $W \times W$ layers with binary activations, and a trainable FP readout $W \to 10$. Every cell trains 25 epochs at batch 128 (3{,}925 steps) with Adam at $3\times10^{-3}$ for the readout (and for the latent weights in the latent arms), flip-lr 1.0, and seeds 0--4. The 8-bit series of Figure~\ref{fig:exchange} and Table~\ref{tab:exchange} use $C{=}1.0$; the naive column of Table~\ref{tab:crule} uses $C{=}1.5$. Dithered Bop keeps an EMA $m \leftarrow (1-\gamma)m + \gamma g$ with $\gamma = 10^{-2}$ at width 192 ($10^{-3}$ in the width sweep), flips when $|m| > \tau$ with the sign of $m$ opposing $w$, resets $m$ to 0 on a flip, and takes the best $\tau \in \{3\times10^{-5}, 10^{-4}, 3\times10^{-4}\}$ at width 192 ($\{10^{-5}, 10^{-4}, 10^{-3}\}$ in the width sweep). Its absmax grid quantizes $m$ onto $2^b$ levels over $[-\max|m|, \max|m|]$ per layer, and the threshold-aligned grid uses $[-\tau, \tau]$ with the flip test $|m| \ge \tau$. The 1-bit-gradient cells round each binary layer's backward operands with per-row absmax scaling after a random-Hadamard transform of block 32 built from a Sylvester matrix and a random sign diagonal, without clipping; the transformer cells of Appendix~\ref{app:extra} clip at $3\times$ the row abs-mean and do not use the transform.

\paragraph{Data.} Shakespeare is the 1{,}115{,}394-byte tiny Shakespeare file with a 65-symbol character vocabulary (or 256 byte values), split 90/10 by position. enwik8 is the first $10^8$ bytes of English Wikipedia, split 90/5/5 by position with the last 5\% unused. wikitext-103 is the raw training split tokenized with the GPT-2 byte-pair encoding into 117.9M tokens, split 90/5/5 by position with the last 5\% unused, so the official validation and test splits are not used.

\paragraph{Compute.} Every run uses one NVIDIA GeForce RTX~4080 SUPER (16\,GB) on a Windows 11 workstation with an AMD Ryzen~5 7600 CPU, PyTorch 2.11 with CUDA 12.8. Wall-clock times per run: MLP 5-seed cells 16--90\,s (longer at width 384 and with 1-bit gradients); Shakespeare 2k steps 0.4--1.4\,min at $d{=}128$, 0.8--2.5\,min at 256, 2.2--6.3\,min at 512; Shakespeare 50k steps 25\,min (FP32, latent) to 40\,min (\ed{}); enwik8 sizes 1.6--3.0, 5.5--11, and 23--55\,min by size; wikitext-103 sizes 8--10, 12--17, and 26--44\,min, with 61--91\,min for the 30k-step cells; 5k-step adaptation 2.5--3.9\,min; a 250-step probe 0.1--11\,min (longest for wikitext-103 at 19M). The CPU kernel runs 1.6 steps per second at 19M parameters (about 21\,min for 2{,}000 steps).

\section{Pre-registered predictions}\label{app:prereg}

Table~\ref{tab:prereg} lists the predictions and pass criteria we wrote down before running the corresponding cells. The records are dated planning and results documents in the public code repository (\texttt{docs/}; Appendix~\ref{app:repro}), not timestamped commits, because the version history begins after the first predictions were recorded. We report every prediction, including the one that failed.

\begin{table}[h]
\centering
\caption{Pre-registered predictions and criteria, in the order they were recorded.}
\label{tab:prereg}
\scriptsize
\setlength{\tabcolsep}{4pt}
\begin{tabular}{p{0.42\linewidth}p{0.2\linewidth}p{0.3\linewidth}}
\toprule
prediction or criterion & record & outcome \\
\midrule
Transformer pass criterion: \ed{}-int8 within $+20\%$ of the latent reference after a flip-lr sweep, else report the negative & \texttt{docs/ADR.md} (ADR-010), 2026-07-14 & passed, $+4.8/+8.1/+11.4\%$ at $d$ 128/256/512 (3 seeds) \\
$d{=}512$ flip-lr optimum at 80--90 (drive $1.05\times10^{-3}$), bracket $\{40, 80, 160\}$, pass if the interior candidate wins & \texttt{docs/scaling-plan.md}, 2026-07-14 & passed, 80 wins (1.782) with 160 tying (1.783) \\
Ladder gate: \ed{} loss runs parallel to the latent reference across sizes, and the two controls (binary activations in both arms, re-probed flip-lrs) were named before running & \texttt{docs/scaling-plan.md}, \texttt{docs/RESULTS\_s2.md} (S2.6), 2026-07-14/15 & passed after the controls, $-2.5/+3.3/+3.9\%$ \\
Threshold-aligned regrid rescues dithered Bop's 4-bit collapse & \texttt{docs/RESULTS\_s2.md} (S2.11), 2026-07-14 to 16 & passed, 46.5 $\to$ 57.0 at \bwt{5} \\
Velocity-scaled drive predicts that heavy-ball momentum cells at flip-lr 1.0 destabilize & \texttt{docs/RESULTS\_s2.md} (S2.4), 2026-07-15 & failed, they did not destabilize (rule does not extend through momentum) \\
Second-order buffer: if 2+2 bits beats a 4-bit buffer the exchange curve improves, else it is dropped & \texttt{docs/ADR.md} (ADR-013), 2026-07-15 & negative, dropped \\
Stall signature: validation loss at 2k above its 1k value identifies stalls, restart with a new seed & \texttt{docs/RESULTS\_s2.md} (S2.9), 2026-07-15 & held on 4 of 4 stalls \\
CPU kernel: warm-started \ed{} reproduces the GPU adaptation loss under the cap, the latent arm cannot run & companion repo \texttt{docs/plan.md}, 2026-07-16 & passed, 1.513 under 275\,MB, latent 325\,MB \\
4-bit 50k cell comparable to 8-bit only at the drive-matched flip-lr (45) & \texttt{docs/RESULTS\_s2.md} (S2.18), 2026-08-23 & 1.681 vs.\ 1.583 \\
\bottomrule
\end{tabular}
\end{table}

\section{Additional results and ablations}\label{app:extra}

\begin{figure}[h]
\centering
\includegraphics[width=0.72\linewidth]{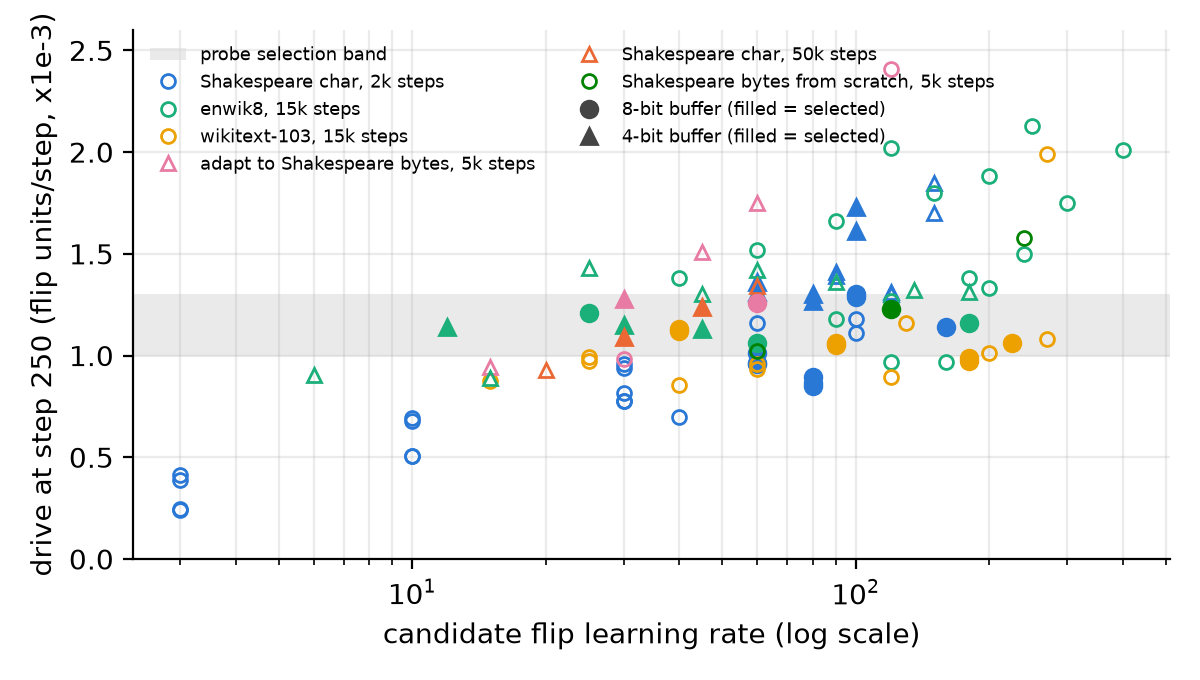}
\caption{Drive at step 250 against the candidate flip-lr for every probe and sweep run behind the transformer cells (three corpora, widths 128--512, 8-bit and 4-bit buffers, from-scratch and warm-started). Filled markers are the flip-lrs selected for the reported cells, and hollow markers are the rejected candidates. The shaded band is the selection rule $[1.0, 1.3]\times10^{-3}$. Every selected flip-lr lies in the band or within 15\% below it, while the selected flip-lrs themselves span 12--225.}
\label{fig:drive}
\end{figure}

\paragraph{Exchange-curve values and drive probes.} Table~\ref{tab:exchange} lists the values plotted in Figure~\ref{fig:exchange}, and Figure~\ref{fig:drive} plots every probe and sweep run behind the transformer cells (\S\ref{sec:drive}).

\begin{table}[h]
\centering
\caption{Test accuracy (\%) on the MLP benchmark by width $W$, mean $\pm$ std over 5 seeds. All \ed{} and Bop series use binary activations and $C{=}1.0$; the FP32+Adam series uses FP activations.}
\label{tab:exchange}
\footnotesize
\setlength{\tabcolsep}{4pt}
\begin{tabular}{lcccccc}
\toprule
$W$ & 64 & 96 & 128 & 192 & 256 & 384 \\
\midrule
latent reference (\bwt{97}) & $40.9 \pm 1.2$ & $47.3 \pm 1.1$ & $52.3 \pm 1.4$ & $57.5 \pm 1.0$ & $59.5 \pm 1.4$ & $62.4 \pm 0.3$ \\
FP32+Adam (\bwt{96}) & $42.8 \pm 1.6$ & $46.9 \pm 1.4$ & $50.3 \pm 1.5$ & $55.8 \pm 1.0$ & $58.3 \pm 0.7$ & $61.2 \pm 0.6$ \\
\ed{}-int8 (\bwt{9}) & $39.8 \pm 1.2$ & $46.0 \pm 2.5$ & $50.9 \pm 1.2$ & $56.3 \pm 1.4$ & $59.4 \pm 0.6$ & $62.3 \pm 0.3$ \\
dithered Bop int8 (\bwt{9}) & $41.3$ & $47.8$ & $51.7$ & $57.2$ & $59.6$ & $62.0$ \\
\ed{}-int4 (\bwt{5}) & $40.3 \pm 1.3$ & $46.8 \pm 1.6$ & $50.4 \pm 1.2$ & $56.4 \pm 1.3$ & $59.8 \pm 0.8$ & $62.5 \pm 0.5$ \\
\bottomrule
\end{tabular}
\end{table}

\paragraph{Controls and budget-split ablations.} Table~\ref{tab:controls} gives the controls of \S\ref{sec:rule} at matched training-state budgets. A $(k{+}1)$-bit saturating integer latent weight with SR updates and sign readout ties the buffer at every matched budget, and a symmetric integer grid satisfies $C{=}1$ by construction, so the buffer coordinates are the frame in which the design variable is visible rather than a separate method. A magnitude-free sign counter \citep{arxiv2412_00119} scores $58.4 \pm 0.8$ at 8 bits, above both \ed{} and the latent reference, but falls to 52.4 at 4 bits and 10.1 at 2 bits, where \ed{} holds 56.4 and 52.3, so below 5 bits per weight only the dithered error-feedback buffer keeps accuracy above the frozen control. A 2-bit buffer plus 2-bit SR-quantized momentum (\bwt{5}) scores 48.9 against 56.4 for the 4-bit buffer, a 4+4 split ties at \bwt{9} (56.3), and a second-order $\Delta\Sigma$ cascade scores 17--29\% at the two budgets tested (\bwt{9} and \bwt{5}), which confirms the buffer-only design. The ECO-style carrier transposes the master-weight-free error feedback of \citet{arxiv2601_22101} to binary weights by decaying the carried residual with a factor $\beta$ each step, so $\beta{=}1$ is the \ed{} buffer.

\begin{table}[h]
\centering
\caption{Controls at width 192 on the MLP benchmark, test accuracy (\%), 5 seeds, binary activations. Budgets are bits of training state per weight.}
\label{tab:controls}
\small
\begin{tabular}{lccc}
\toprule
method & \bwt{9} & \bwt{5} & \bwt{3} \\
\midrule
\ed{} buffer ($C{=}1.0$) & 56.3 & 56.4 & 52.3 \\
saturating integer latent weight, SR updates, sign readout & 56.2 & 56.1 & 52.1 \\
magnitude-free sign counter \citep{arxiv2412_00119} & $58.4 \pm 0.8$ & 52.4 & 10.1 \\
buffer plus SR-quantized momentum (budget split) & 56.3 & 48.9 & \\
second-order $\Delta\Sigma$ cascade & 29.1 & 22.5 & -- \\
\midrule
ECO-style decayed carrier, $\beta = 0.9 / 0.99 / 1$ & \multicolumn{3}{c}{45.4 / 55.3 / 56.3 at 8 buffer bits} \\
\bottomrule
\end{tabular}
\end{table}

\paragraph{One-bit gradients.} Unbiased 1-bit SR of the backward operands, with per-row absmax scaling and with or without random-Hadamard mixing at block sizes 32--128, fails on transformer gradients at every learning rate, with validation loss never falling below the unigram level. Clip-then-SR (clip at $3\times$ the row abs-mean) restores learning, and mixing adds nothing. Scoped to the weight-gradient path \citep{arxiv2407_11722} with a per-layer adaptive drive, the 1-bit-gradient configuration reaches $2.017 \pm 0.009$ (3 seeds) at $d{=}256$ and 2k steps, within 8\% of FP-gradient \ed{} (1.881 at a constant flip-lr 60). Table~\ref{tab:interaction} gives the interaction with buffer width on the MLP benchmark.

\begin{table}[h]
\centering
\caption{Cost of 1-bit gradients (SR after a random-Hadamard transform of block 32, per-row absmax, no clipping) by buffer width at width 192 on the MLP benchmark, test accuracy (\%), 5 seeds, binary activations, $C{=}1.0$.}
\label{tab:interaction}
\small
\begin{tabular}{lccc}
\toprule
buffer bits & FP gradients & 1-bit gradients & cost (points) \\
\midrule
8 & $56.3 \pm 1.4$ & $55.2 \pm 1.3$ & $-1.0$ \\
4 & $56.4 \pm 1.3$ & $48.6 \pm 1.0$ & $-7.7$ \\
2 & $52.3 \pm 1.2$ & $38.1 \pm 0.9$ & $-14.2$ \\
\bottomrule
\end{tabular}
\end{table}

\section{Diagnostics behind \S\ref{sec:cap}}\label{app:cap}

Cell for the stall study: enwik8, 6L/512, int8 buffer, flip-lr 180 with linear decay, binary activations, evaluations every 100 steps.

\paragraph{Cohorts.} Zero-initialized buffers, seeds 0--9: 2 stalls by the registered rule (seeds 2 and 3, S2.9) plus one transient spike that the rule does not catch. Buffers initialized uniformly in $[-0.5, 0.5]$, seeds 0--9: 3 stalls (two late collapses at step 3k and one at 2k) and one slow run. A diagnostic replicate of all 20 launches: 4 avalanches above threshold (flip rate $\times \geq 1.4$ and first-block pre-activation scale $\times \geq 1.3$ between consecutive 100-step evaluations) plus one minor event; block 0 was the epicenter every time. Twelve continuations from a re-created step-2000 checkpoint with identical weights, buffers, Adam state, and random streams all ran clean, and CUDA non-associativity alone spread their best loss by 0.016--0.018 nats.

\begin{figure}[h]
\centering
\includegraphics[width=0.9\linewidth]{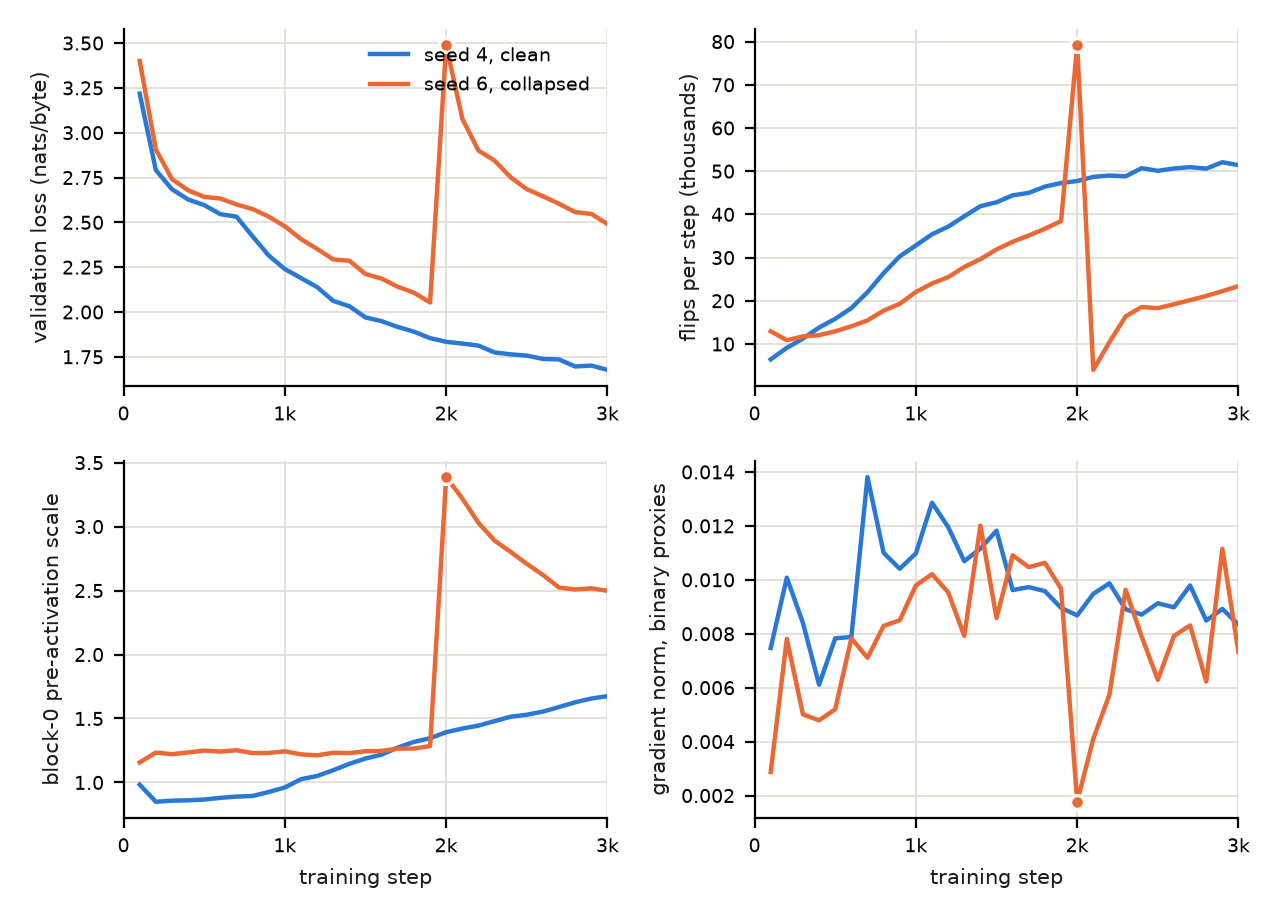}
\caption{The same two launches at 100-step resolution: validation loss, flips per step, first-block pre-activation scale, and the gradient norm reaching the binary proxies. The marker is the collapsed launch at step 2000. Everything that defines the event happens inside one evaluation window; recovery runs at a tenth of the normal flip rate.}
\label{fig:avalanche}
\end{figure}

\paragraph{One captured event.} Table~\ref{tab:event} and Figure~\ref{fig:avalanche} show the diagnostics of one collapsed launch (initialization 0.5, seed 6) next to a clean one from the same cohort (seed 4).

\begin{table}[h]
\centering
\caption{The captured avalanche, initialization 0.5, seed 6: diagnostics at the 100-step evaluations around the event. The FP-remnant columns (LayerNorm gains, head norm, remnant gradient) stay on their prior trend while the binary core's columns jump.}
\label{tab:event}
\scriptsize
\setlength{\tabcolsep}{2.5pt}
\begin{tabular}{rrrrrrrrr}
\toprule
step & val loss & flips/step & buffer $\bar{|e|}$ & act.\ scale, blocks 0 / 1 / 2 & grad, binary proxies & grad, FP remnant & LN gain mean / max & head norm \\
\midrule
1800 & 2.107 & 36.7k & 0.449 & 1.26 / 0.87 / 0.73 & 0.0106 & 0.81 & 1.012 / 1.16 & 19.4 \\
1900 & 2.053 & 38.4k & 0.455 & 1.28 / 0.89 / 0.74 & 0.0097 & 0.72 & 1.013 / 1.16 & 19.6 \\
\textbf{2000} & \textbf{3.488} & \textbf{79.2k} & \textbf{0.535} & \textbf{3.39 / 2.58 / 1.29} & \textbf{0.0018} & 0.45 & 1.014 / 1.17 & 19.7 \\
2100 & 3.078 & 4.0k & 0.530 & 3.22 / 2.59 / 1.31 & 0.0041 & 0.55 & 1.014 / 1.18 & 19.6 \\
2500 & 2.685 & 18.3k & 0.520 & 2.71 / 2.38 / 1.31 & 0.0063 & 0.57 & 1.013 / 1.18 & 20.4 \\
3000 & 2.490 & 23.4k & 0.516 & 2.50 / 2.02 / 1.22 & 0.0073 & 0.69 & 1.012 / 1.17 & 21.0 \\
\bottomrule
\end{tabular}
\end{table}

\paragraph{Capped cohorts.} Zero initialization, seeds 0--9, 3000 steps: 0 avalanches, validation loss at 3k $1.687 \pm 0.020$ against $1.829 \pm 0.113$ uncapped ($1.767$ for the seven clean uncapped runs). Full 15k runs, seeds 0--2: $1.2664$, $1.2754$, $1.2834$, no stall. On wikitext-103 at 19M the capped arm (remnant Adam $3\times10^{-4}$, flip-lr 225) ran monotonically: 5.79, 5.40, 5.18, 5.04, 4.93, 4.90, 4.78, 4.73, 4.70, 4.66, 4.63, 4.59, 4.56, 4.53, 4.50 at every 1k steps, against 6.21, 5.89, 6.29, 5.83, 5.59, 5.41, 6.55, 5.78, 5.38, 5.61, 5.27, 5.16, 5.06, 4.97, 4.92 uncapped.

\paragraph{Remnant controls at 19M on wikitext-103} (15k steps, seed 0 unless stated). Frozen token and position embeddings: the latent arm 5.122, the \ed{} arm diverged in 3 of 3 launches. Remnant Adam $10^{-4}$: latent 4.625, \ed{} 4.610 (one of five launches) and 4.615 with the cap. Remnant Adam decaying $3\times10^{-4} \to 3\times10^{-5}$: \ed{} 4.681 (one of three launches). Remnant Adam $3\times10^{-5}$: latent 5.143, \ed{} diverged in 3 of 3 launches at its probe-set flip-lr 400. Embeddings and head frozen (LayerNorms trainable): latent 7.243, \ed{} 7.279.

\paragraph{Bop on the transformer} (enwik8, 4L/256, binary activations, 15k steps, 3 seeds, $\pm$ std). Bop needs its published scale here, $\gamma = 10^{-4}$ and $\tau = 10^{-7}$ (bracketed: $\tau = 3\times10^{-8}$ gives 1.627, $3\times10^{-7}$ gives 1.686; $\gamma = 10^{-3}$ oscillates or freezes), and the 2k-step cells of Table~\ref{tab:gpt} are too short for its EMA to settle. A frozen core (no flips) reaches 2.069. Bop with the FP32 EMA (\bwt{33}): $1.524 \pm 0.011$; \ed{}-int8 (\bwt{9}): $1.516 \pm 0.012$; Bop with a 4-bit EMA on the absmax grid: $2.463 \pm 0.240$, two of three launches early-stopped with about one flip per step after step 3k; on the $\tau$-aligned grid: $1.669 \pm 0.004$; \ed{}-int4: $1.707 \pm 0.003$. The rescue is partial here where it was total on the MLP: the transformer's EMA values ($\sim 4\times10^{-8}$) sit on the lowest levels of a 16-level grid over $[-\tau, \tau]$, where stochastic rounding adds flips that are partly noise.

\paragraph{4-bit buffer on wikitext-103} (probe-set flip-lrs 20 / 45 / 45). Uncapped: 2L/128 diverged at 7k and 3k, third launch 5.224; 4L/256 diverged at 7k, 5k, 2k; 6L/512 completed at the first launch, 4.749; 6L/512 at 30k diverged at 5k in all three launches. Capped: 5.275, 4.974, 4.742, each at the first launch; at 30k the first launch diverged at 7k and the second completed at 4.681. On enwik8 the capped 4-bit cells (flip-lrs 12 / 30 / 45) are 2.007, 1.711, 1.509 against 2.002, 1.707, 1.510 uncapped.

\section{Reproducibility}\label{app:repro}

All experiments are seeded, with a single CPU generator per run and a documented consumption order; CUDA kernel non-associativity still separates reruns of one configuration by up to about 0.02 nats in stable runs, and can decide whether an uncapped launch stalls (Appendix~\ref{app:cap}). Every result row is one JSON Lines (JSONL) record that carries its configuration (the latent-arm rows of the adaptation log omit the Adam learning rate, which the run script sets). Code, run scripts, logs, and the dated records are public at \url{https://github.com/pierreolivierbonin/error-diffusion-binary-weight-lms} under the MIT License. The \texttt{main} branch holds the experiments of Sections~3--5, and each experiment behind \S\ref{sec:cap} has its own branch with the exact code that produced its logs; Table~\ref{tab:repro} maps each reported artifact to its script and log. A large language model (LLM) assisted with drafting, experiment orchestration, and literature search under the author's direction. Seeded code produced all results. All citations were checked programmatically against arXiv, DOI, and Semantic Scholar records. The author reviewed and takes responsibility for all content.

\begin{table}[h]
\centering
\scriptsize
\setlength{\tabcolsep}{4pt}
\begin{tabular}{p{0.3\linewidth}p{0.36\linewidth}p{0.27\linewidth}}
\toprule
Artifact & Script(s) & Log(s) \\
\midrule
Fig.~\ref{fig:concept} & \texttt{tools/plot\_halftone\_concept\_sim.py} & self-contained, fixed seed \\
MLP headline, Fig.~\ref{fig:exchange} & \texttt{headline.sh}, \texttt{widthsweep.sh}, \texttt{widthsweep\_int4c1.sh}, \texttt{r4\_fixes.sh} & \texttt{headline\_5seed}, \texttt{widthsweep\_5seed}, \texttt{widthsweep\_int4c1}, \texttt{r4\_widthsweep\_int8c1} \\
Table~\ref{tab:crule} (\ed{}) & \texttt{p2clamp.sh}, \texttt{p2boundary.sh} & \texttt{p2clamp}, \texttt{p2boundary\_5seed} \\
Table~\ref{tab:crule} (Bop) & \texttt{s2bopalign.sh}, \texttt{s2bopalign2.sh} & \texttt{s2bopalign}, \texttt{s2bopalign2} \\
Integer latent, counter, ECO-style carrier & \texttt{s2intlat.sh}, \texttt{s2baselines.sh} & \texttt{s2intlat}, \texttt{s2baselines} \\
Second-order buffer & \texttt{s2sigma.sh} & \texttt{s2sigma} \\
Table~\ref{tab:gpt} & \texttt{p3gpt.sh}, \texttt{audit\_fix.sh}, \texttt{s2int4gpt.sh}, \texttt{r4\_fixes.sh}, 50k cells & \texttt{p3gpt}, \texttt{audit\_fix}, \texttt{s2int4gpt}, \texttt{r4\_gpt\_seed2}, \texttt{s2converged}, \texttt{s2decay}, \texttt{s2stationarity} \\
1-bit gradients, interaction (App.~\ref{app:extra}) & \texttt{p3qfix.sh}, scoped runs, \texttt{audit\_fix.sh}, \texttt{r4\_fixes.sh} & \texttt{p3qfix}, \texttt{p3scope}, \texttt{audit\_fix\_mlp}, \texttt{r4\_mlp} \\
Table~\ref{tab:ladder} (byte ladder) & \texttt{s2controls.sh}, \texttt{s2ladderseeds.sh}, \texttt{s2int4gpt3.sh} & \texttt{s2controls}, \texttt{s2ladderseeds}, \texttt{s2int4gpt3} \\
Table~\ref{tab:ladder} (tokenized ladder) & \texttt{s2tokladder.sh} & \texttt{s2tokladder}, \texttt{s2tok512long} \\
Deploy-then-adapt (\S\ref{sec:adapt}) & \texttt{s2adapt1.sh}, \texttt{s2adapt2.sh}, \texttt{s2int4gpt2.sh}, \texttt{r4\_fixes.sh} & \texttt{s2adapt}, \texttt{s2int4gpt}, \texttt{r4\_adapt} \\
Drive rule (\S\ref{sec:drive}, Fig.~\ref{fig:drive}) & probes, \texttt{s2tokprobe.sh}, \texttt{s2int4probe*.sh} & \texttt{s2*probe*}, \texttt{parityprobe}, \texttt{s2rule} \\
\S\ref{sec:cap}, Appendix~\ref{app:cap} & \texttt{r5\_e2*.sh}, \texttt{r5\_e3*.sh}, \texttt{r5\_e5*.sh}, \texttt{r5\_e1\_int4\_drift.sh}, \texttt{r5\_e4*.sh} & \texttt{r5\_e2*}, \texttt{r5\_e3*}, \texttt{r5\_e5*}, \texttt{r5\_e1\_int4\_drift\_v2}, \texttt{r5\_e4*} \\
\bottomrule
\end{tabular}
\caption{Scripts (\texttt{scripts/}, \texttt{tools/}) and JSONL logs (\texttt{results/}) behind each artifact. The files of the last row are on the repository branches whose names start with \texttt{r5-e1} to \texttt{r5-e5}.}
\label{tab:repro}
\end{table}

\end{document}